\documentclass[lettersize,journal]{IEEEtran}
\usepackage{amsmath,amsfonts}
\usepackage{algorithmic}
\usepackage{algorithm}
\usepackage{array}
\usepackage[caption=false,font=normalsize,labelfont=sf,textfont=sf]{subfig}
\usepackage{textcomp}
\usepackage{stfloats}
\usepackage{url}
\usepackage{verbatim}
\usepackage{graphicx}
\usepackage{cite}
\usepackage{algorithm}
\usepackage{algorithmic}
\usepackage{multirow} 
\usepackage{color}
\usepackage{bbding}
\usepackage{pifont}
\usepackage{makecell}
\usepackage{subcaption} 
\usepackage{colortbl}
\usepackage[table]{xcolor}

\usepackage{graphicx}
\usepackage{mathrsfs}
\usepackage{tikz}
\usepackage{amsmath}
\usepackage{bbding}
\usepackage{hyperref}

\newcommand{\circled}[2][0.2mm]{\tikz[baseline=-0.57ex]{
    \node[shape=circle, draw, line width=#1, inner sep=#1, minimum size=7pt, font=\scriptsize, align=center] (char) {#2};
}}

\begin{document}

\title{DepthFusion: Depth-Aware Hybrid Feature Fusion for LiDAR-Camera 3D Object Detection}

\author{Mingqian Ji, Jian Yang, Shanshan Zhang$^{\dag}$
\thanks{$^{\dag}$ indicates corresponding author: Shanshan Zhang.}%

\thanks{Mingqian Ji, Jian Yang, and Shanshan Zhang are with the PCA Lab, Key Lab of Intelligent Perception and Systems for High-Dimensional Information of Ministry of Education, and Jiangsu Key Lab of Image and Video Understanding for Social Security, School of Computer Science and Engineering, Nanjing University of Science and Technology (e-mail: mingqianji@njust.edu.cn; csjyang@mail.njust.edu.cn; shanshan.zhang@njust.edu.cn).}
}

\markboth{Journal of \LaTeX\ Class Files,~Vol.~14, No.~8, August~2021}%
{Shell \MakeLowercase{\textit{et al.}}: A Sample Article Using IEEEtran.cls for IEEE Journals}


\maketitle

\begin{abstract}
State-of-the-art LiDAR-camera 3D object detectors usually focus on feature fusion. However, they neglect the factor of depth while designing the fusion strategy. In this work, we are the first to observe that different modalities play different roles as depth varies via statistical analysis and visualization. Based on this finding, we propose a Depth-Aware Hybrid Feature Fusion (DepthFusion) strategy that guides the weights of point cloud and RGB image modalities by introducing depth encoding at both global and local levels. Specifically, the Depth-GFusion module adaptively adjusts the weights of image Bird's-Eye-View (BEV) features in multi-modal global features via depth encoding. Furthermore, to compensate for the information lost when transferring raw features to the BEV space, we propose a Depth-LFusion module, which adaptively adjusts the weights of original voxel features and multi-view image features in multi-modal local features via depth encoding. Extensive experiments on the nuScenes and KITTI datasets demonstrate that our DepthFusion method surpasses previous state-of-the-art methods. Moreover, our DepthFusion is more robust to various kinds of corruptions, outperforming previous methods on the nuScenes-C dataset.
\end{abstract}

    
\section{Introduction}
\IEEEPARstart{3}{D} object detection has a wide range of applications in the fields of autonomous driving and robotics. A large number of previous works have successfully focused on using a single modality, such as point cloud or images, to design efficient 3D object detectors. However, the performance of these detectors reaches a bottleneck due to the limitations of modality characteristics. For instance, the point cloud modality can only provide rich geometric information while lacking detailed semantic information; the image modality can only provide rich texture information while lacking three-dimensional spatial information. To address the aforementioned issues, we are highly motivated to obtain comprehensive information that represents objects by designing a LiDAR-camera 3D object detector.

\begin{figure}[t]
      \centering
      \includegraphics[width=0.95 \linewidth]{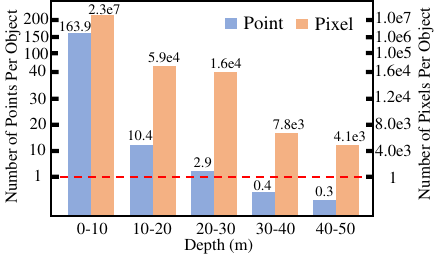}
      \caption{Statistical analysis on the nuScenes-mini dataset. The average numbers of points and pixels for each object at different depths.}
      \label{vis1}
    \end{figure}

\begin{figure}[t]
      \centering
      \includegraphics[width=1.0 \linewidth]{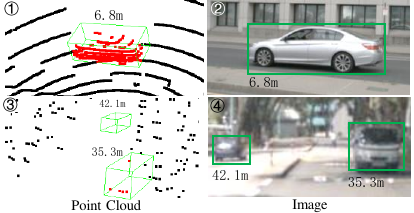}
      \caption{Visualization analysis on the nuScenes-mini dataset. Examples of near-range and long-range objects in images and point cloud. Points within the bounding boxes are colored \textcolor{red}{red} for observation.}
      \label{vis2}
    \end{figure}

In recent years, LiDAR-camera 3D object detection develops rapidly. Some works propose effective methods to integrate information from two modalities at the feature level. However, they all overlook an important factor of depth in their fusion strategies. To understand how point cloud and image information vary with depth, we first conduct statistical and visualization analysis on the nuScenes-mini dataset \cite{caesar2020nuscenes}, a comprehensive and diverse dataset designed for autonomous driving, and find that: (1) The number of points representing objects at near range is relatively large, which allows us to accurately determine the object's location, size, and category, even without the aid of images. As shown in Fig. \ref{vis1}, there is an average of 163.7 points per object within 0-10 meters, which is a substantial number. We also visualize a car at 6.8 meters in Fig. \ref{vis2} \ding{172} and find it encompasses a considerable number of points, well representing the shape.
In contrast, some background noise in the image may interfere with detection (Fig. \ref{vis2} \ding{173}). (2) As the depth increases, the number of points representing objects decreases rapidly. As shown in Fig. \ref{vis1}, the number of points within 30-50 meters falls below one per object, meaning that many objects are even not represented by any points, such as the object at 42.1 meters in Fig. \ref{vis2} \ding{174}. In contrast, the complete objects may still be observed on the image, as in Fig. \ref{vis2} \ding{175}, where the image information becomes more important. To address the above problems, we propose a feature fusion strategy that adaptively adjusts the importance of the two modalities based on depth.

Specifically, we propose a novel method for LiDAR-camera 3D object detection, namely Depth-Aware Hybrid Feature Fusion (DepthFusion). The innovation lies in adaptively adjusting the weights of features by introducing depth encoding to hybrid feature fusion at both global and local levels. The fusion strategy consists of two crucial components: Depth-GFusion (DGF) module and Depth-LFusion (DLF) module. In DGF, we take point cloud Bird's-Eye-View (BEV) features and image BEV features as inputs, and dynamically adjust the weights of image BEV features based on depth during fusion by utilizing a global feature fusion with a depth embedding. To compensate for the information lost when transforming raw features to BEV space, we enhance the fused BEV features at a lower cost by utilizing the original instance features. In DLF, we obtain 3D boxes by utilizing a Region Proposal Network (RPN). Then, the 3D boxes are projected into both LiDAR voxel features and multi-view image features to crop out corresponding local instance features with more detailed information. Afterward, we take these as inputs and dynamically adjust the weights of local multi-view image features and local LiDAR voxel features based on depth through the use of a local feature fusion with the depth embedding. In the end, we update local features for each object on the global feature map to enhance the detailed instance information of multi-modal global features for detection.
Our contributions are summarized as follows:

\begin{itemize}
    \item We are the first to observe that depth is an important factor to consider while fusing LiDAR point cloud features and RGB image features for 3D object detection. From our statistical and visualization analysis, we can see that image features play different roles as depth varies.
    
    \item We propose a depth-aware hybrid feature fusion strategy that dynamically adjusts the weights of features during feature fusion by introducing depth encoding at both global and local levels. The above strategy can obtain high-quality features for detection, fully leveraging the advantages of different modalities at various depths.

    \item Our method is evaluated on the nuScenes \cite{caesar2020nuscenes} dataset, KITTI \cite{geiger2012kitti} dataset, and a more challenging nuScenes-C \cite{dong2023nuscenes-c} dataset, outperforming previous LiDAR-camera methods and being robust to various kinds of data corruptions.
\end{itemize}

\section{Related Work}
Since our method is based on conducting 3D object detection using data from multiple modalities, including point cloud and images, we briefly review recent works in the following fields: LiDAR-based 3D object detection, camera-based 3D object detection, and LiDAR-camera 3D object detection.

\subsection{LiDAR-based 3D Object Detection}
LiDAR-based 3D object detectors only take the point cloud as input. Based on their different data representations, they can be divided into point-based, voxel-based, and point-voxel-based methods. The feature extraction networks of point-based methods typically extract features directly from the point cloud through a point-based backbone, such as PointRCNN \cite{shi2019pointrcnn}. The voxel-based methods first convert the point cloud into voxels and then extract voxel features through a 3D sparse convolution network, such as VoxelNet \cite{zhou2018voxelnet} and SP-Det \cite{an2023sp}. Point-voxel-based methods like PV-RCNN \cite{shi2020pv} combine the above two methods to extract and fuse point and voxel features. The purpose of these approaches is to capture the geometric spatial information of the point cloud. However, point cloud is sparse and incomplete, lacking detailed texture information, which greatly limits the detection performance.

\subsection{Camera-based 3D Object Detection}
Camera-based 3D object detectors only take images as inputs. Depending on the form of inputs, they can be divided into monocular, stereo, and multi-view 3D object detectors. Early works like FCOS3D \cite{wang2021fcos3d} input a monocular image and utilize 2D object detectors to directly predict 3D bounding boxes, but these approaches have limited capability in capturing spatial information. Subsequently, stereo and multi-view 3D object detectors are proposed to obtain more precise depth information by constructing spatial relationships among multiple images, such as BEVDet \cite{huang2021bevdet}. These methods successfully achieve purely visual 3D object detection, but they do not perform as well as LiDAR-based methods, because the spatial depth information provided by images is not as direct and precise as that provided by point cloud.

\subsection{LiDAR-Camera 3D Object Detection}
LiDAR-camera 3D object detectors take point cloud and images as inputs, and can be classified into early-fusion-based, intermediate-fusion-based, and late-fusion-based 3D object detectors based on the location of multi-modal information fusion. 

Early-fusion-based methods perform at the point level, where the typical approach involves enhancing the raw point cloud with semantic information extracted from images. PointPainting \cite{vora2020pointpainting} decorates the raw point cloud with semantic scores from 2D semantic segmentation. 
Similarly, PointAugmenting \cite{wang2021pointaugmenting}, VirConv \cite{wu2023virtual} and VirPNet \cite{wang2024virpnet} enhance the raw point cloud using virtual point features extracted from a 2D semantic segmentation network. However, early-fusion-based methods are sensitive to alignment errors between the two modalities.

Intermediate-fusion-based methods perform at the feature level. Transfusion \cite{bai2022transfusion} first proposes to utilize the transformer for fine-grained fusion from LiDAR BEV features and multi-view image features. VPFNet \cite{zhu2022vpfnet} uses the virtual point features extracted from stereo images to enhance point cloud feature. FUTR3D \cite{chen2023futr3d} encodes each modality using deformable attention \cite{zhu2020deformable} in its own coordinate and concatenates them for fusion. 3D-DFM \cite{lin20223d} proposes a dynamic learnable filters to adaptively interact images with point features based on an anchor-free architecture. CL3D \cite{lin2022cl3d} employs a two-stream design with point-guided fusion and an IoU-aware head to tackle the misalignment between localization and classification. BEVFusion \cite{liang2022bevfusion,liu2023bevfusion} projects both point cloud and images to BEV space for BEV feature fusion. SparseFusion \cite{xie2023sparsefusion} extracts instance-level features from both two modalities separately, and fuses them to perform detection. Similarly, ObjectFusion \cite{cai2023objectfusion} utilizes 3D proposals from LiDAR modality to extract instance-level features for fusion. CMT \cite{yan2023cmt} proposes the simultaneous interaction between the object queries and multi-modal features in the transformer encoder and decoder. LoGoNet \cite{li2023logonet} and IS-Fusion \cite{yin2024isfusion} propose feature fusion at both the instance level and scene level. The intermediate-fusion-based methods gradually become a mainstream approach due to the diversity of fusion strategies.

Late-fusion-based methods perform at the bounding box level. Typically, CLOCs \cite{pang2020clocs} obtains 2D and 3D bounding boxes by separately using 2D and 3D object detectors, and then combines them to achieve more accurate 3D bounding boxes. However, the interaction between modalities in late-fusion-based methods is very limited, which constrains model performance.

These LiDAR-camera methods successfully outperform single-modal methods. However, their feature fusion methods do not take depth into account. In contrast, our approach introduces depth information to guide the hybrid feature fusion,  boosting the performance of the detector.

\section{Methodology}

In this section, we first give an overview of our proposed LiDAR-camera 3D object detector, and then provide a detailed introduction to our proposed feature fusion method. 

\subsection{Overview} \label{overview}

    \begin{figure*}[t]
      \centering
       \includegraphics[width= \linewidth]{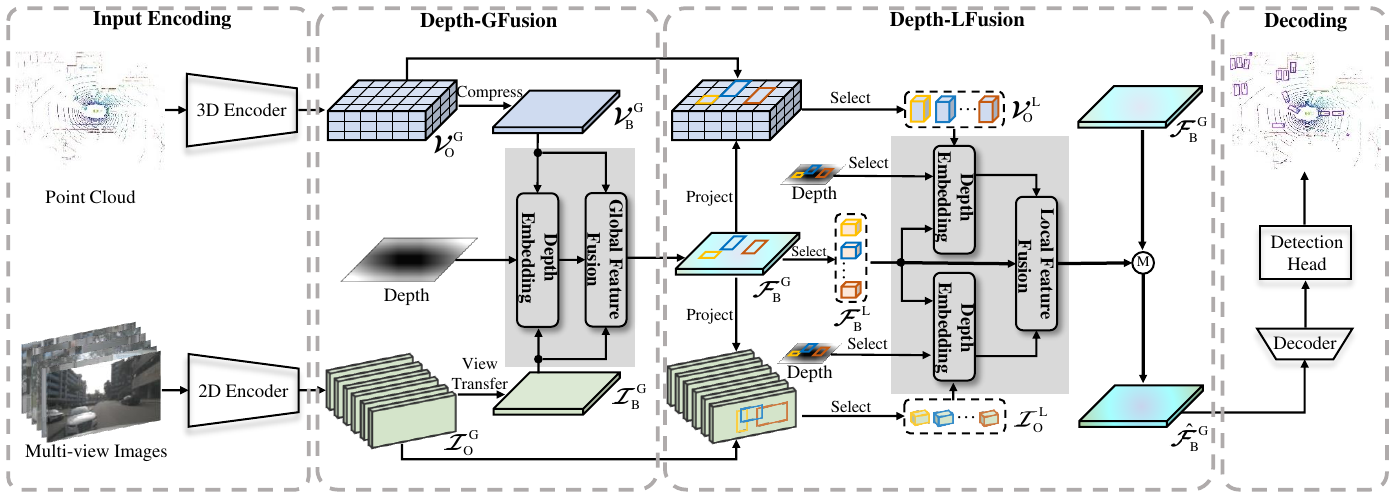}
       \caption{Overview of our method. It introduces depth encoding in both global and local feature fusion to obtain depth-adaptive multi-modal representations for detection. \circled[0.01em]{\tiny M} is the merge operation. }
      \label{pipeline}
    \end{figure*}

LiDAR and camera sensors provide complementary information at different depth ranges. LiDAR delivers highly accurate and dense geometric information in close-range areas (e.g., within 30 meters). However, beyond 30 meters, the point cloud becomes extremely sparse (fewer than one point per object), making it difficult to capture object geometry reliably. In contrast, cameras provide high-resolution semantic and texture information that remains dense even at long range, though they lack direct depth measurements. To exploit the strengths of both modalities, we propose a LiDAR-camera 3D object detection method via depth-aware hybrid feature fusion. As illustrated in Fig. \ref{pipeline}, our approach consists of two important feature fusion modules: Depth-GFusion and Depth-LFusion. In the following, we briefly describe the detection pipeline.

\textbf{Inputs.} First, we take the point cloud $P$ and multi-view images $I$ as inputs, where the point cloud consists of a set of points: $P=\{P_1, P_2, \cdots, P_{N_l}\}$, and each point has four dimensions: X-axis, Y-axis, Z-axis, and intensity; the multi-view images comprise $N_c$ images: $I=\{I_1, I_2, \cdots, I_{N_c}\}$, each image captured by its corresponding camera.

\textbf{Input Encoding.} For the point cloud $P$, we use a 3D encoder to extract raw global voxel features $\mathcal{V}_O^G$; for the multi-view images $I$, we use a 2D encoder to extract image features of all views $\mathcal{I}_O^G$. 

\textbf{Hybrid Feature Fusion.} Then, for voxel features $\mathcal{V}_O^G$, we compress the height dimension to obtain point cloud BEV features $\mathcal{V}_B^G$; for image features $\mathcal{I}_O^G$, following BEVFusion \cite{liu2023bevfusion}, we adopt a bin-based projection method to transform perspective-view features into the bird's eye view. Specifically, pixel-wise depth distributions are predicted and used to project the image features into BEV space, resulting in the image BEV feature $\mathcal{I}_B^G$. To fully leverage the features from two modalities, we design a DGF module that aims to dynamically adjust the weights of image BEV features based on depth values during feature fusion. Please refer to Sec. \ref{DGF} for more details. To compensate for the information lost when transforming raw features to BEV space, we propose a DLF module that, based on depth, utilizes the raw features to enhance the detailed information of each object instance in global multi-modal features. It consists of three processes: local feature selection, local feature fusion, and merging local features into global features. First, we obtain the local multi-modal BEV features $\mathcal{F}_B^L$, local voxel features $\mathcal{V}_O^L$, and local multi-view image features $\mathcal{I}_O^L,$ by cropping the corresponding global features based on the 3D boxes obtained from an RPN; then, it dynamically and individually adjusts the weights of each local feature of $\mathcal{V}_O^L$ and $\mathcal{I}_O^L$ based on depth values during feature fusion; finally, we update local features for each object on the global feature map. Please refer to Sec. \ref{DLF} for more details. In this way, we obtain enhanced multi-modal global features for detection. 

\textbf{Decoding.} Based on the enhanced multi-modal global features $\hat{\mathcal{F}}_B^G$ that contain rich semantic and spatial information, we utilize a transformer decoder and a detection head to predict the object categories and 3D bounding boxes.

\subsection{Depth-GFusion} \label{DGF}

\begin{figure}[t]
      \centering
      \includegraphics[width=0.95 \linewidth]{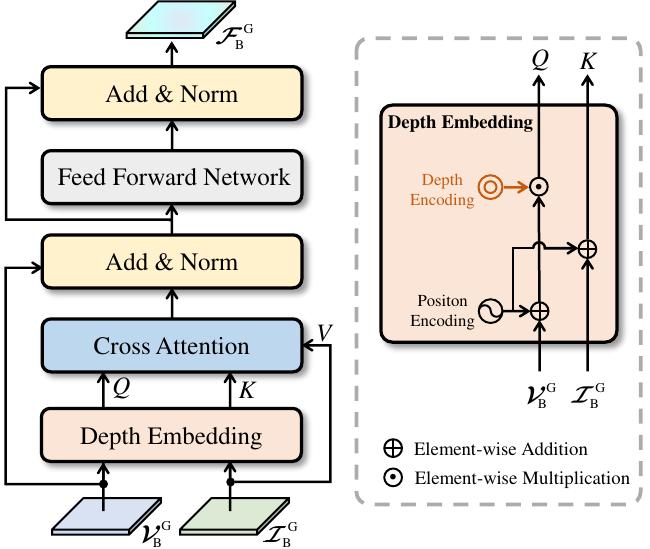}
      \caption{Illustration of the DGF.}
      \label{DGF_figure}
    \end{figure}

As shown in Fig. \ref{DGF_figure}, the DGF module consists of a global feature fusion with a depth embedding. In the following, we provide a detailed explanation of each component.

\subsubsection{Depth Encoding} \label{depth encoder}
We introduce depth encoding $D$ in feature fusion to dynamically adjust the weights of image BEV features during fusion. First, we build a depth matrix $M$ to store the depth value of each position element $p_k$ represented as: 
    \begin{equation}
     p_k=\{ (x_k, y_k): d_k \}, k\in[1, n],
        \label{equation_Z}
    \end{equation}
where $(x_k, y_k)$ are the positional coordinates, $d_k$ is the depth value, and $n$ is the number of elements. Then, we use Euclidean distance to calculate the distance between every element's spatial location $(x_k, y_k)$ and the ego coordinate element's location $(x_\frac{n}{2}, y_\frac{n}{2})$:
    \begin{equation}
        d_k=E((x_k, y_k), (x_\frac{n}{2}, y_\frac{n}{2})), k \in [1, n],
        \label{m_k}
    \end{equation}
where we denote $E(\cdot)$ as the Euclidean distance calculation. The resulting depth matrix $M$ is implemented as a dictionary, where each key is a BEV coordinate $(x_k, y_k)$ and the value is its corresponding precomputed depth $d_k$. To avoid redundant computation, the depth matrix $M$ serves as a lookup table. In our fusion design, its usage differs across modules: In DGF, which fuses global BEV features, the entire matrix $M$ is utilized directly in its full form without dictionary lookup. In DLF, which performs instance-level fusion, the lookup table allows efficient retrieval of depth values for selected BEV elements, significantly reducing the computational overhead during localized fusion. Since the size of the BEV features is large and the depth distribution is simple, to avoid introducing additional parameters, the depth encoding $D$ is obtained by applying sine and cosine functions \cite{vaswani2017attention} to the depth matrix.

\subsubsection{Global Feature Fusion} \label{global-fusion transformer}
In the depth embedding, we take the point cloud BEV features $\mathcal{V}_B^G \in \mathbb{R}^{W \times H \times C}$ and image BEV features $\mathcal{I}_B^G \in \mathbb{R}^{W \times H \times C}$ as inputs, and the positional encoding $P$ is added to the original BEV features through element-wise addition to represent the spatial location of each element in the BEV grid, which ensures that the model is aware of the relative positions of the features within the BEV space. Then, we integrate depth encoding $D$ by multiplying it with the point cloud BEV features, as this allows the $query$ to capture the depth-based weight variations when querying the image BEV features, reflecting how the importance of image features changes with depth. The image BEV features are then queried as the corresponding $key$ and $value$. We utilize the multi-head cross attention to achieve the interacted feature $\hat{\mathcal{V}}_B^G$ based on depth:
    \begin{equation}
        \hat{\mathcal{V}}_B^G = \text{softmax}\left[\frac{((\mathcal{V}_B^G + P) \cdot D) (\mathcal{I}_B^G + P)^T}{\sqrt{C}}\right] \mathcal{I}_B^G,
    \end{equation}
where $\mathcal{V}_B^G$ and $\mathcal{I}_B^G$ are the point cloud and image BEV features; $P$ and $D$ are the positional encoding and our proposed depth encoding; $C$ is the dimension of the above features. Afterward, we aggregate the information from both modalities to obtain the fused features $\mathcal{F}_B^G$:
    \begin{equation}
        \mathcal{F}_B^G = N(FFN(N(\hat{\mathcal{V}}_B^G + \mathcal{V}_B^G)) + N(\hat{\mathcal{V}}_B^G + \mathcal{V}_B^G)),
    \end{equation}
where $N(\cdot)$ is a normalization layer; $FFN(\cdot)$ specifies a feed-forward network containing two convolution operations. In this way, we obtain fused features in which the image features play different roles as the depth varies. 

\subsection{Depth-LFusion} \label{DLF}
The DLF module is designed to enhance the instance-level BEV features by incorporating both LiDAR BEV instance features and the original image instance features into the BEV space, which consists of a local feature selection and a local feauture fusion with a depth embedding. Through this depth-aware local fusion, the module dynamically adjusts the fusion weights based on instance depth and effectively enriches the BEV representation with complementary modality information. In the following, we provide a detailed explanation of each component.

\subsubsection{Local Feature Selection}

To compensate for the information lost when transforming point cloud features and image features to BEV space, we enhance the instance details of fused BEV features $\mathcal{F}_B^G$ using instance features from raw voxel features $\mathcal{V}_O^G$ and multi-view image features $\mathcal{I}_O^G$. Specifically, we utilize an RPN to regress $t$ 3D boxes based on the BEV features $\mathcal{F}_B^G$. We directly crop the global fused BEV features $\mathcal{F}_B^G$ based on the regressed 3D boxes to obtain the local fused BEV features $\mathcal{F}_B^L \in \mathbb{R}^{c \times t} $. On the other hand, we project the 3D boxes onto the raw voxel features and multi-view image features to obtain their corresponding local features before global fusion, preserving richer information for each object instance. Specifically, we utilize the voxel pooling operation \cite{deng2021voxel}, followed by a 3D convolution operation and a linear layer, to extract local voxel features $\mathcal{V}_O^L \in \mathbb{R}^{c \times t} $; we transform the 3D boxes from bird's eye view to perspective view, and utilize the RoI Align operation \cite{he2017mask}, followed by a linear layer, to extract instance image features $\mathcal{I}_O^L \in \mathbb{R}^{c \times t}$. By doing this, we obtain the hybrid (before \& after global fusion) local features, which will be sent to the subsequent fusion module.

\subsubsection{Local Feature Fusion} \label{local-fusion transformer}
In the local feature fusion, the weights of each local raw feature are dynamically adjusted based on depth values during feature fusion, and we update local features for each object on the global feature map. Specifically, we take the local multi-modal BEV features $\mathcal{F}_B^L$, local voxel features $\mathcal{V}_O^L$, and local multi-view image features $\mathcal{I}_O^L$ as inputs, and integrate the positional encoding and our proposed depth encoding to the local multi-modal BEV features, forming the query ${Q_{\mathcal{F}}^L}$. The local multi-view image features and local voxel features are respectively queried as the corresponding key $K_{\mathcal{I}}^L$, $K_{\mathcal{V}}^L$ and value $V_{\mathcal{I}}^L$, $V_{\mathcal{V}}^L$.
The two multi-head cross-attention modules are utilized to achieve the interacted features $\hat{Q}_\mathcal{F}^L, \hat{Q}{_\mathcal{F}^L}'$. Note that the computation process of multi-head cross attention is similar to that described in Sec. \ref{global-fusion transformer} and is omitted here. Afterward, we aggregate the above features:
    \begin{equation}
        \hat{\mathcal{F}}_B^L = FFN(Cat(\hat{Q}_\mathcal{F}^L + \mathcal{F}_B^L, \hat{Q}{_\mathcal{F}^L}' + \mathcal{F}{_B^L}')),
    \end{equation}
where $Cat(\cdot)$ is the concatenation operation; $FFN(\cdot)$ is used to extract high-dimensional, fine-grained local features containing two linear layers. As a result, we obtain enhanced local features by dynamically calling back rich information in raw modalities at various depths. Afterward, we update the global features $\mathcal{F}_B^G$ by inserting the enhanced local features at corresponding locations.
    \begin{figure}[t]
      \centering
      \includegraphics[width=0.95 \linewidth]{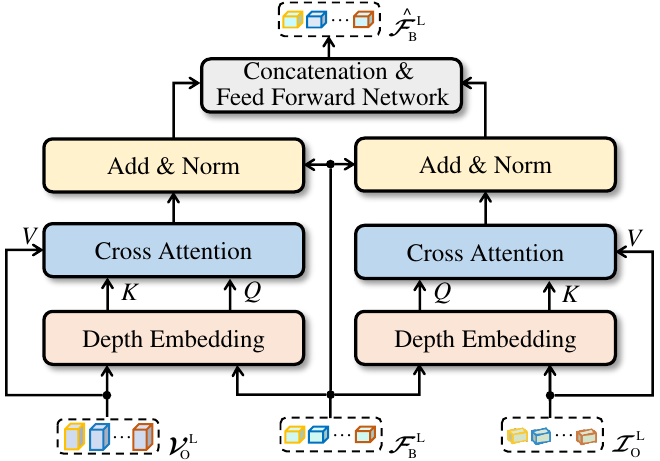}
      \caption{Illustration of the DLF.}
      \label{DLF_figure}
    \end{figure}
\section{Experiments}
In this section, we will first introduce the dataset and evaluation metrics. Then, we will compare our method with the state-of-the-art methods on nuScenes, KITTI, and nuScenes-C datasets. Moreover, we provide additional comparative experiments, including comparisons at different depths, comparisons on small-sized objects, and comparisons of model efficiency. Finally, we will show the ablation studies and visualization. 

\subsection{Experimental Setup}
\textbf{Datasets and evaluation metrics.} We evaluate our proposed DepthFusion on the nuScenes \cite{caesar2020nuscenes}, KITTI \cite{geiger2012kitti} datasets, and a more challenging dataset of nuScenes-C \cite{dong2023nuscenes-c} with data corruptions. The nuScenes dataset provides 700 scene sequences for training, 150 scene sequences for validation, and 150 scene sequences for testing. 
The KITTI dataset contains 7481 samples for training and 7518 samples for testing. We split the original training set into training and validation sets, following \cite{shi2020pv}. 
The nuScenes-C dataset provides 27 corruptions with 5 severities on the nuScenes validation set, including corruptions at the weather, sensor, motion, object, and alignment levels. For the nuScenes and nuScenes-C datasets, we use the nuScenes detection scores (NDS) and mean Average Precision (mAP) to evaluate our detection results, where NDS is a comprehensive metric that combines object translation, scale, orientation, velocity, and attribute errors. For the KITTI dataset, we use 3D Average Precision (3D AP) with recall 40 positions to evaluate our detection results.

\textbf{Implementation details.} We use BEVFusion \cite{liu2023bevfusion} as the baseline method, and implement the proposed DepthFusion under the open-source framework MMDetection3D \cite{mmdet3d2020}. Specifically, on the nuScenes dataset, for the LiDAR branch, we use VoxelNet \cite{zhou2018voxelnet} with FPN \cite{yan2018second} as the 3D encoder. The voxel size is set to [0.075m, 0.075m, 0.1m], and the range of point cloud is [-54m, 54m] along the X-axis, [-54m, 54m] along the Y-axis, and [-3m, 5m] along the Z-axis. For the image branch, we use the ResNet18 \cite{he2016resnet}, ResNet50 \cite{he2016resnet}, SwinTiny \cite{liu2022convnet}, and ConvNeXtS \cite{liu2022convnet} with FPN \cite{lin2017fpn} as the 2D image encoder of DepthFusion-light, -base, -large, and -huge, respectively. Correspondingly, the resolution of input images is resized to 256 $\times$ 704, 320 $\times$ 800, 384 $\times$ 1056, and 900 $\times$ 1600. On the KITTI dataset, for the LiDAR branch, the voxel size is set to [0.05m, 0.05m, 0.1m], and the range of point cloud is [0, 70.4m] along the X-axis, [-40m, 40m] along the Y-axis, and [-3m, 1m] along the Z-axis. For the image branch, we follow \cite{li2023logonet} to use SwinTiny with FPN as the 2D image encoder, and the resolution of input images is resized to 187 $\times$ 621. Additionally, we utilize BEVPoolV2 \cite{huang2022bevpoolv2} to obtain image BEV features. Following \cite{liu2023bevfusion}, the feature size $W \times H$ is set to 180 $\times$ 180, the channel $C$ is set to 128, and the channel $c$ is also set to 128. Following \cite{xie2023sparsefusion}, the number of regressed 3D boxes $t$ is set to 200. More implementation details are provided in supplementary materials.


    \begin{table*}[t!]
        \centering
        \caption{Comparisons with the state of the art on the nuScenes $\tt{validation}$ and $\tt{test}$ sets. FPS is measured on a 3090 GPU by default, and * denotes the inference speed on an A100 GPU referred from the original paper. Note that all results are obtained without any model ensemble or test time augmentation.}
        \setlength{\tabcolsep}{1.3mm}{
        \scalebox{1.0}{
        \begin{tabular}{c | c|c|c|c c|c c} 
        \hline
        \multirow{2}{1.5cm}{\ Methods} & \multirow{2}{1.8cm}{\quad Present at} & \multirow{2}{3.8cm}{\quad Image Size - 2D Backbone} & \multirow{2}{0.6cm}{\ FPS} & \multicolumn{2}{c}{Validation} & \multicolumn{2}{|c}{Test} \\
        & & & & NDS & mAP & NDS & mAP \\
        \hline
        \hline
        \multicolumn{8}{c}{Image Backbone: ResNet50\cite{he2016resnet}} \\
        \hline
        Transfusion \cite{bai2022transfusion} & CVPR'22 & 320 $\times$ 800-ResNet50 & 6.5 & 71.3 & 67.5 & 71.7 & 68.9 \\
        DeepInteraction \cite{yang2022deepinteraction} & NeurIPS'22 & 448 $\times$ 800-ResNet50 & 1.9 & 72.4 & 69.9 & 73.4 & 70.8 \\
        MSMDFusion \cite{jiao2023msmdfusion} & CVPR'23 & 448 $\times$ 800- ResNet50 & 2.1 & 72.1 & 69.7 & 74.0 & 71.5 \\
        FocalFormer3D \cite{chen2023focalformer3d} & ICCV'23 & 320 $\times$ 800-ResNet50 & 9.2* &  73.1 & 70.1 & 73.9 & 71.6 \\
        VirPNet \cite{wang2024virpnet} & TMM'24 & 320 $\times$ 800-ResNet50 & - & 73.2 & 70.4 & - & - \\
        \hline
        \rowcolor{gray!20}
        \textbf{DepthFusion-base (Ours)} & - & 320 $\times$ 800-ResNet50 & 8.7 & \textbf{74.0} & \textbf{71.2} & \textbf{74.7} & \textbf{71.7} \\
        \hline
        \hline
        \multicolumn{8}{c}{Image Backbone: SwinTiny\cite{liu2021swintransformer}} \\
        \hline
        BEVFusion \cite{liang2022bevfusion} & NeurIPS'22 & 448 $\times$ 800-SwinTiny & 0.7* & 71.0 & 67.9 & 71.8 & 69.2 \\
        BEVFusion \cite{liu2023bevfusion} & ICRA'23 & 256 $\times$ 704- SwinTiny & 9.6 & 71.4 & 68.5 & 72.9 & 70.2 \\
        ObjectFusion \cite{cai2023objectfusion} & ICCV'23 & 256 $\times$ 704- SwinTiny & - & 72.3 & 69.8 & 73.3 & 71.0 \\
        SparseFusion \cite{xie2023sparsefusion} & ICCV'23 & 256 $\times$ 704- SwinTiny & 4.4 & 72.8 & 70.5 & 73.8 & 72.0 \\
        IS-Fusion \cite{yin2024isfusion} & CVPR'24 & 384 $\times$ 1056-SwinTiny & 3.2* & 74.0 & 72.8 & 75.2 & 73.0 \\
        GAFusion \cite{li2024gafusion} & CVPR'24 & 448 $\times$ 800-SwinTiny & - & 73.5 & 72.1 & 74.9 & 73.6 \\
        \hline
        \multicolumn{8}{c}{Image Backbone: Others} \\
        \hline
        AutoAlignV2 \cite{chen2022AutoAlignV2} & ECCV'22 & 640 $\times$ 1280-CSPNet \cite{wang2020cspnet} & 4.8* & 71.2 & 67.1 & 72.4 & 68.4 \\
        UVTR \cite{li2022unifying} & NeurIPS'22 & 640 $\times$ 1280-ResNet101 \cite{he2016resnet} & 1.8 & 70.2 & 65.4 & 71.1 & 67.1 \\
        FUTR3D \cite{chen2023futr3d} & CVPR'23 & 900 $\times$ 1600-VOVNet \cite{lee2019vovnet} & 3.3* & 68.0 & 64.2 & 72.1 & 69.4 \\
        UniTR \cite{wang2023unitr} & ICCV'23 & 256 $\times$ 704-DSVT \cite{wang2023dsvt} & 9.3* & 73.3 & 70.5 & 74.5 & 70.9 \\
        CMT \cite{yan2023cmt} & ICCV'23 & 640 $\times$ 1600-VOVNet & 6.0* & & 70.3 & 74.1 & 72.0 \\
        UniPAD \cite{yang2024unipad} & CVPR'24 & 900 $\times$ 1600-ConvNeXtS \cite{liu2022convnet} & - & 73.2 & 69.9 & 73.9 & 71.0 \\
        SparseLIF \cite{zhang2024sparselif} & ECCV'24 & 900 $\times$ 1600-VOVNet & - & 74.6 & 71.2 & - & \- \\
        \hline
        \rowcolor{gray!20}
        \textbf{DepthFusion-light (Ours)} & - & 256 $\times$ 704-ResNet18 & \textbf{13.8} & 73.3 & 69.8 & 74.2 & 70.9 \\
        \rowcolor{gray!20}
        \textbf{DepthFusion-large (Ours)} & - & 384 $\times$ 1056-SwinTiny & 5.7 & 74.4 & 72.3 & 75.4 & 72.8 \\
        \rowcolor{gray!20}
        \textbf{DepthFusion-huge (Ours)} & - & 900 $\times$ 1600-ConvNeXtS & 3.4 & \textbf{74.9} & \textbf{72.9} & \textbf{75.8} & \textbf{73.6} \\
        \hline
        \end{tabular}
            }
        }
        \label{Comparison--sota}
    \end{table*}
    
     \begin{table*}[t!]
        \centering
        \caption{Comparisons with the state of the art on the KITTI $val$ set. The results are on \textbf{3D AP} with IoU=0.7, 0.5, 0.5 for three classes: Car, Pedestrian, and Cyclist. We use bold for the best results, and underline for the second best results.}
        \setlength{\tabcolsep}{1.3mm}{
        \scalebox{1.0}{
        \begin{tabular}{c | c c c c|c c c c|c c c c| c} 
        \hline
        \multirow{2}{1.5cm}{\ \ Methods} & \multicolumn{4}{c}{Car} & \multicolumn{4}{|c}{Pedestrian} & \multicolumn{4}{|c|}{Cyclist} & \multirow{2}{0.8cm}{\ mAP} \\
        \cline{2-13}
        & Easy & Mod. & Hard & mAP & Easy & Mod. & Hard & mAP & Easy & Mod. & Hard & mAP \\
        \hline
        PointFusion \cite{xu2018pointfusion} & 77.9 & 63.0 & 53.3 & 64.7 & 33.4 & 28.0 & 23.4 & 28.3 & 49.3 & 29.4 & 27.0 & 35.3 & 42.8 \\
        CLOCs \cite{pang2020clocs} & 89.5 & 79.3 & 77.4 & 82.1 & 62.9 & 56.2 & 50.1 & 56.4 & 87.6 & 67.9 & 63.7 & 73.1 & 70.5 \\
        EPNet \cite{huang2020epnet} & 88.8 & 78.7 & 78.3 & 81.9 & 66.7 & 59.3 & 54.8 & 60.3 & 83.9 & 65.6 & 62.7 & 70.7 & 71.0 \\
        CAT-Det \cite{zhang2022cat} & 90.1 & 81.5 & 79.2 & 83.6 & \underline{74.1} & 66.4 & 58.9 & 66.5 & 87.6 & 72.8 & 68.2 & 76.2 & 75.4 \\
        3D-DFM \cite{lin20223d} & 91.9 & 84.9 & 82.5 & 86.4 & - & - & - & - & - & - & - & - & - \\
        VPFNet \cite{zhu2022vpfnet} & 91.0 & 83.2 & 78.2 & 84.1 & - & - & - & - & - & - & - & - & - \\
        CL3D \cite{lin2022cl3d} & 90.3 & 83.2 & 78.9 & 84.1 & - & - & - & - & - & - & - & - & - \\
        VirPNet \cite{wang2024virpnet} & 92.0 & 80.8 & 78.9 & 83.9 & 72.1 & \textbf{67.6} & \underline{60.6} & 66.7 & 90.8 & \underline{75.1} & \underline{70.9} & \underline{79.1} & 76.6 \\
        FocalsConv \cite{chen2022focal} & 92.3 & 85.3 & 83.0 & 86.8 & - & - & - & - & - & - & - & - & - \\
        VFF \cite{li2022voxel} & 92.3 & 85.5 & 82.9 & 86.9 & 73.3 & 65.1 & 60.0 & 66.1 & 89.4 & 73.1 & 69.9 & 77.5 & 76.9 \\
        LoGoNet \cite{li2023logonet} & 92.0 & 85.0 & 84.3 & 87.1 & 70.2 & 63.7 & 59.5 & 64.5 & \underline{91.7} & \textbf{75.4} & \textbf{72.4} & \textbf{79.8} & 77.1 \\
        VirConv \cite{wu2023virtual} & \underline{94.9} & \textbf{90.0} & \textbf{88.1} & \underline{91.0} & 73.3 & 66.9 & 60.4 & \underline{66.9} & 90.0 & 73.9 & 69.1 & 77.7 & 78.5 \\
        \hline
        \rowcolor{gray!20}
        \textbf{DepthFusion (Ours)} & \textbf{95.6} & \underline{89.6} & \underline{87.7} & \textbf{91.0} & \textbf{75.6} & \underline{67.5} & \textbf{63.7} & \textbf{68.9} & \textbf{91.8} & 74.5 & 69.8 & 78.7 & \textbf{79.5} \\
        \hline
        \end{tabular}
            }
        }
        \label{Comparison--sota kitti}
    \end{table*}

\subsection{Comparisons on the nuScenes Dataset}

Aiming for a fair comparison, we categorize previous methods based on the types of 2D backbones into ResNet50-based, SwinTiny-based, and others, and provide four versions of our proposed method, named DepthFusion-light, DepthFusion-base, DepthFusion-large, and DepthFusion-huge. The comparison results are shown in Tab. \ref{Comparison--sota}. (1) Compared with the ResNet50-based methods, our DepthFusion-base outperforms the top methods, FocalFormer3D \cite{chen2023focalformer3d} and VirPNet \cite{wang2024virpnet}, by up to 1 pp w.r.t. NDS under the same configuration. Specifically, we reach 74.0\% NDS and 71.2\% mAP on the validation set, and 74.7\% NDS and 71.7\% mAP on the test set, while maintaining a comparable inference speed of 8.7 FPS on a 3090 GPU. (2) Moreover, our DepthFusion-light surpasses the typical BEVFusion \cite{liu2023bevfusion} by up to 1 pp w.r.t. all metrics using a lighter 2D backbone, and achieves a real-time inference speed of 13.8 FPS. (3) Compared with the SwinTiny-based methods and others, our DepthFusion-large outperforms the top method IS-Fusion \cite{yin2024isfusion} under the same configuration, and runs 2x faster than it. In addition, our DepthFusion-huge with a larger image size and 2D backbone surpasses all previous SOTA methods. Specifically, we reach 74.9\% NDS on the validation set, and 75.8\% NDS on the test set, while achieving the inference speed of 3.4 FPS on a 3090 GPU. Overall, our method achieves higher detection accuracy and faster inference speed.

\subsection{Comparisons on the KITTI Dataset}

 To further evaluate the effectiveness of our proposed method, we evaluate our DepthFusion on the KITTI dataset. Tab. \ref{Comparison--sota kitti} shows the results on the KITTI $val$ set. Compared with previous multimodal methods, DepthFusion achieves state-of-the-art mAP performance, surpassing the top method VirConv \cite{wu2023virtual} by 1 pp. The results demonstrate the effectiveness of our method on the KITTI dataset. Notably, for the small-sized pedestrian class, our DepthFusion achieves 63.7\% AP at the hard difficulty level and 68.9\% mAP, outperforming VirConv by up to 2 pp. This indicates that our method is also effective for small-sized objects, while they are at a distant range or extremely sparse. Overall, these results further validate the effectiveness of DepthFusion.

    \begin{table*}[t!]
        \centering
        \noindent
            \centering
            \caption{Robustness experiments on nuScenes-C. Numbers are \textbf{NDS} / \textbf{mAP}.}
            \setlength{\tabcolsep}{1.4mm}{
            \scalebox{1.0}{
            \begin{tabular}{c | c || c c c c c | c} 
            \hline
            \multirow{2}{1.2cm}{Methods} & \multicolumn{6}{c|}{Corruption} & \multirow{2}{1.0cm}{Average}  \\
            \cline{2-7}
            & None & Weather & Sensor & Motion & Object & Alignment \\
            \hline
            FUTR3D \cite{chen2023futr3d} & 68.05 / 64.17 & 62.75 / 55.51 & 63.66 / 56.83 & 53.16 / 44.43 & 65.45 / 61.04 & 62.83 / 57.60 & 62.82\textcolor{red}{$^{\downarrow5.23}$}  / 56.99\textcolor{red}{$^{\downarrow7.18}$}  \\
            TransFusion \cite{bai2022transfusion} & 69.82 / 66.38& 65.42 / 59.37 & 66.17 / 59.82 & 51.52 / 41.47 & 68.28 / 64.38 & 61.98 / 54.94 & 63.74\textcolor{red}{$^{\downarrow6.08}$}  / 58.73\textcolor{red}{$^{\downarrow7.65}$} \\
            BEVFusion \cite{liu2023bevfusion} & 71.40 / 68.45 & 67.54 / 61.87 & 67.59 / 61.80 & 55.19 / 47.30 & 68.01 / 65.14 & 63.94 / 58.71 & 66.06\textcolor{red}{$^{\downarrow5.34}$}  / 61.03\textcolor{red}{$^{\downarrow7.42}$}  \\
            \hline
            \rowcolor{gray!20}
            \textbf{DepthFusion-light (Ours)} & \textbf{73.30} / \textbf{69.75} & \textbf{72.19} / \textbf{67.48} & \textbf{69.16} / \textbf{62.87} & \textbf{57.07} / \textbf{47.52} & \textbf{71.01} / \textbf{67.11} & \textbf{67.24} / \textbf{62.38} & \textbf{68.67}\textcolor{red}{$^{\downarrow4.63}$}  / \textbf{63.07}\textcolor{red}{$^{\downarrow6.68}$} \\
            \hline
            \end{tabular}
                }
            }
            \label{Robustness-NDS}
    \end{table*}

    \begin{table*}[t!]
        \centering
        \caption{Comparisons on the nuScenes dataset for different object types at a far distance ($>$30m). The numbers are \textbf{AP}.}
        \label{far distance comparison}
        \setlength{\tabcolsep}{1.75mm}{
        \scalebox{1.0}{
        \begin{tabular}{ c| c c c c c c c c c c} 
        \hline
        Methods & Car & Truck & Bus & Trailer & Construction\_Vehicle & Pedestrian & Motorcycle & Bicycle & Traffic\_Cone & Barrier \\
        \hline
        TransFusion-L \cite{bai2022transfusion} & 70.3 & 43.2 & 55.0 & 27.8 & 14.2 & 37.4 & 35.3 & 22.7 & 0.0 & 0.0  \\
        BEVFusion \cite{liu2023bevfusion} & 72.0 & 49.9 & 57.2 & 29.5 & 14.5 & 53.9 & 39.3 & 24.5 & 0.0 & 0.0 \\
        \rowcolor{gray!20}
        \hline
        \textbf{DepthFusion-light (Ours)} & \textbf{77.2} & \textbf{54.8} & \textbf{64.5} & \textbf{39.8} & \textbf{17.3} & \textbf{74.7} & \textbf{47.9} & \textbf{35.8} & \textbf{23.5} & \textbf{39.6} \\
        \hline
        \end{tabular}
        }
        }
    \end{table*}

    \begin{table}[t!]
        \centering
        \caption{Comparisons on the nuScenes validation set at different depths. The numbers are \textbf{mAP}.}
        \setlength{\tabcolsep}{1.2mm}{
        \scalebox{0.97}{
        \begin{tabular}{c | c c c} 
        \hline
        Methods & Near (0-20m) & Middle (20-30m) & Far ($>$30m) \\
        \hline
        TransFusion-L \cite{bai2022transfusion} & 77.5 & 60.9 & 30.5 \\
        BEVFusion \cite{liu2023bevfusion} & 79.4 & 64.9 & 34.1  \\
        \hline
        \rowcolor{gray!20}
        \textbf{DepthFusion-light (Ours)} & \textbf{80.3} & \textbf{66.5} & \textbf{47.5} \\
        \hline
        \end{tabular}
            }
        }
        \label{Table Evaluation at Different Depths}
    \end{table}
    
\subsection{Robustness to Corruptions}

We further implement some experiments on the nuScenes-C \cite{dong2023nuscenes-c} dataset to evaluate the model's robustness under various corruptions, including changes in weather, data loss or temporal-spatial misalignment in multi-modal inputs, etc. The results for different kinds of corruptions are shown in Tab. \ref{Robustness-NDS}, and more detailed results for each fine-grained corruption are shown in supplementary materials. We find that our DepthFusion-light still achieves an average performance of 68.67\% NDS and 63.07\% mAP under various corruptions, which only decreases by 4.63 pp w.r.t. NDS and 6.68 pp w.r.t. mAP, compared to its performance without corruptions. Performance drop is smaller than that observed with previous methods including BEVFusion \cite{liu2023bevfusion} across all kinds of corruptions, indicating that our DepthFusion-light possesses superior robustness. Furthermore, we observe that our DepthFusion-light is particularly robust against weather and object corruptions, where the performance drop is less than 3 pp. The more stable performance indicates that our method is more friendly to practical applications, where data corruption may occur.

\subsection{Comparisons at Different Depths}
Since our fusion strategy is depth-aware, it is necessary to validate our method at different depths. On the nuScenes dataset, we categorize annotation and prediction ego distances into three groups: Near (0-20m), Middle (20-30m), and Far ($>$30m). As shown in Tab. \ref{Table Evaluation at Different Depths}, compared to BEVFusion \cite{liang2022bevfusion}, our DepthFusion-light consistently improves performance across all depth ranges. Specifically, our method achieves a 47.5\% mAP at a far distance ($>$30m), surpassing ObjectFusion by 13.4 pp. These results indicate that our method is more effective across different depths, especially in detecting distant objects.

\subsection{Comparisons for Different Objects at a Far Distance}
To further demonstrate the effectiveness of our method for distant object detection, we conduct experiments on the nuScenes dataset to evaluate its performance across various object categories at far distances. As shown in Tab. \ref{far distance comparison}, our DepthFusion-light consistently outperforms both TransFusion-L \cite{bai2022transfusion} and BEVFusion \cite{liu2023bevfusion} across all categories. Notably, DepthFusion-light achieves significant improvements in detecting normal-sized objects such as cars and trucks. Moreover, our method successfully detects smaller objects, including traffic cones and barriers, whereas the comparison methods fail to detect these categories. These results underscore the effectiveness of our depth-aware feature fusion approach, which markedly enhances the detection of distant objects in challenging scenarios.

    \begin{table}[t!]
        \centering
        \caption{Robustness under LiDAR signal drop-out ($>$40m) on the nuScenes validation set. The numbers are \textbf{mAP}.}
        \setlength{\tabcolsep}{2.6mm}{
        \scalebox{0.96}{
        \begin{tabular}{c | c c} 
        \hline
        Methods & Natural Drop-out & Simulated Drop-out \\
        \hline
        TransFusion-L \cite{bai2022transfusion} & 15.4 & - \\
        BEVFusion \cite{liu2023bevfusion} & 20.4 & 16.9\textcolor{red}{$^{\downarrow3.5}$} \\
        \hline
        \rowcolor{gray!20}
        \textbf{DepthFusion-light (Ours)} & \textbf{33.6} & \textbf{31.2}\textcolor{red}{$^{\downarrow2.4}$}  \\
        \hline
        \end{tabular}
            }
        }
        \label{Table robustness under far-distance LiDAR signal drop-out}
    \end{table}

\subsection{Robustness under Far-Distance LiDAR Signal Drop-out}
To evaluate the robustness of our method under far-distance sensor degradation, we conduct experiments on the nuScenes validation set with two kinds of LiDAR signal drop-out scenarios: \textit{Natural Drop-out} reflects real-world cases where LiDAR points beyond 40 meters are sparse or missing due to occlusion or sensor limitations; and \textit{Simulated Drop-out} reflects all LiDAR points beyond 40 meters are removed during inference to mimic extreme signal loss. As shown in Table \ref{Table robustness under far-distance LiDAR signal drop-out}, our DepthFusion-light our DepthFusion-light achieves 33.6\% mAP under natural drop-out, significantly outperforming BEVFusion (20.4\% mAP). Under simulated drop-out, our method still maintains strong performance with 31.2\% mAP, exhibiting only a slight drop of 2.4 pp. In contrast, BEVFusion suffers a larger decrease of 3.5 pp. These results demonstrate that our depth-aware cross-modal fusion effectively compensates for far-distance LiDAR signal loss.

    \begin{figure*}[t]
        \centering
        \includegraphics[width=0.95 \linewidth]{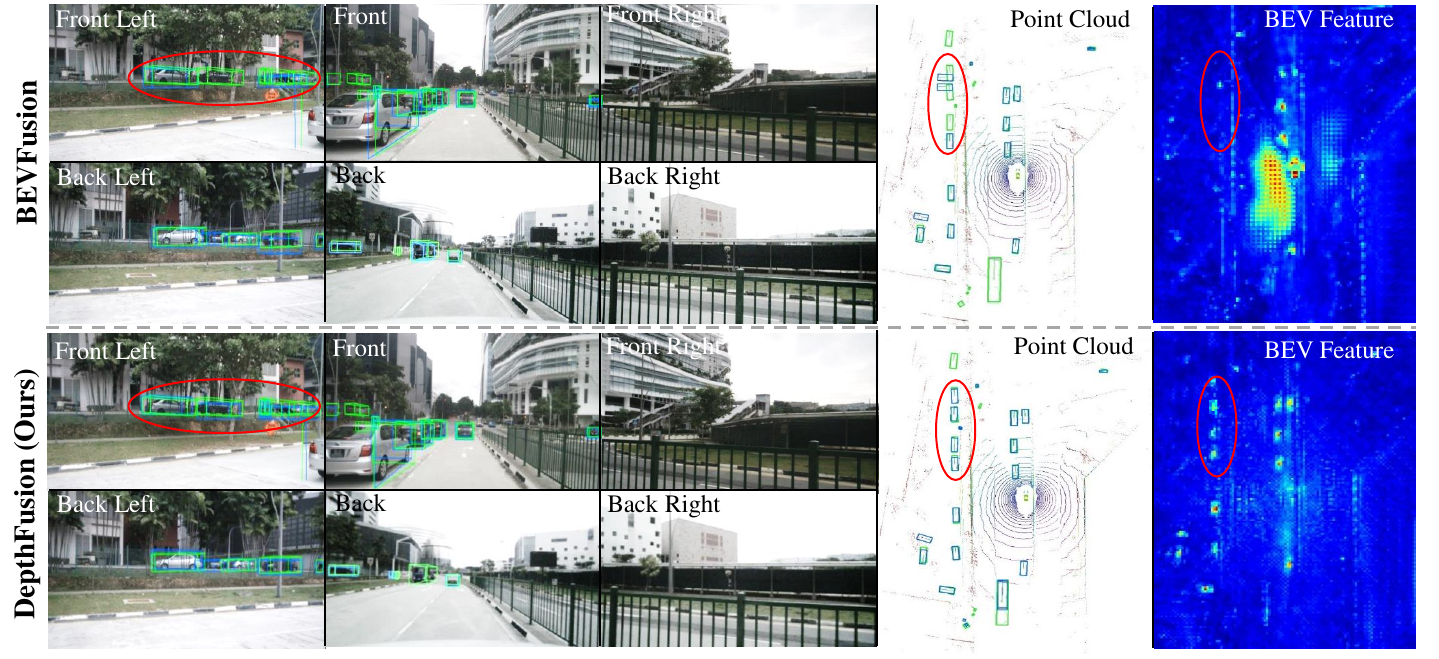}
        \caption{Qualitative detection results and BEV features of BEVFusion and ours. We show the ground truth boxes in \textcolor{green}{green}, and the prediction boxes in \textcolor{blue}{blue}. We use \textcolor{red}{red} circles to highlight the comparisons of ours with BEVFusion.}
        \label{vis_figure}
    \end{figure*}

    \begin{table}[t!]
    \centering
        \caption{Comparisons of parameter, FLOPs, and latency.}
        \label{Computational Efficiency}
        \centering
        \setlength{\tabcolsep}{0.3mm}{
        \scalebox{1.0}{
        \begin{tabular}{ c| c c | c c c} 
        \hline
        Methods & NDS & mAP & Parameters (M) & FLOPs (G) & Latency (ms) \\
        \hline
        BEVFusion & 71.4 & 68.5 & 40.84 & 253.2 & 104.17 \\
        BEVFusion + Ours & 73.8 & 71.5 & 56.95 & 271.7 & 126.51 \\
        \hline
        DepthFusion-light & 73.3 & 69.8 & 40.38 & 242.6 & 72.46\\
        DepthFusion-base &  74.0 & 71.2 & 56.15 & 822.8 & 114.94 \\
        DepthFusion-large & 74.4 & 72.3 & 56.94 & 1508.2 & 175.44 \\
        DepthFusion-huge & 74.9 & 72.9 & 78.85 & 1973.8 & 294.12 \\
        \hline
        \end{tabular}
            }
        }
    \end{table}

    \begin{table}[t!]
        \centering
        \caption{Ablation studies of each proposed module.}
        \centering
        \setlength{\tabcolsep}{2.6mm}{
        \scalebox{1.0}{
        \begin{tabular}{ c | c c} 
        \hline
         Methods & NDS & mAP \\
        \hline
         Baseline & 71.4 & 68.5 \\
         Baseline $\rightarrow$ + DGF & \quad \ \ 72.4\textcolor{green}{$^{\uparrow1.0}$} & \quad \ \ 69.4\textcolor{green}{$^{\uparrow0.9}$}  \\
         Baseline $\rightarrow$ + DLF & \quad \ \ 72.7\textcolor{green}{$^{\uparrow1.3}$} &  \quad \ \ 69.3\textcolor{green}{$^{\uparrow0.8}$} \\
         \hline
         Baseline $\rightarrow$ + DGF $\rightarrow$ + DLF & \quad \ \ \textbf{73.3}\textcolor{green}{$^{\uparrow1.9}$} & \quad \ \ \textbf{69.8}\textcolor{green}{$^{\uparrow1.3}$} \\
         Baseline $\rightarrow$ + DLF $\rightarrow$ + DGF & \quad \ \ 73.0\textcolor{green}{$^{\uparrow1.6}$} & \quad \ \ 69.4\textcolor{green}{$^{\uparrow0.9}$} \\
        \hline
        \end{tabular}
            }
        }
        \label{ablation-each module}
    \end{table}

\subsection{Model Computational Efficiency}
\begin{figure}[t]
        \centering
        \includegraphics[width=1.0 \linewidth]{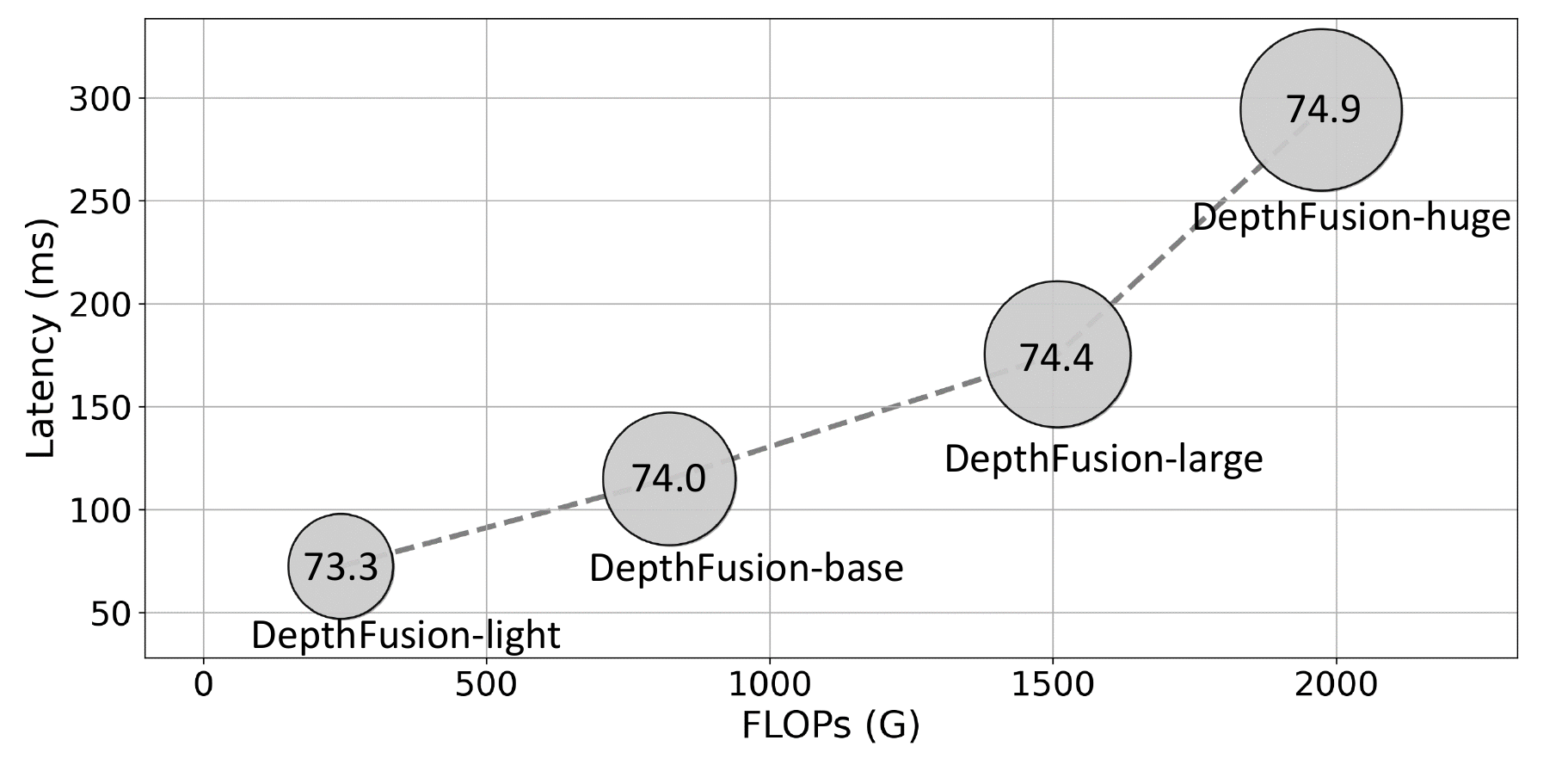}
        \caption{Trade-off between performance (NDS) and computational cost (FLOPs and latency) across different DepthFusion variants. Circle size indicates the \textbf{NDS}, with larger circles indicating better performance.}
        \label{trade-off}
    \end{figure}

We provide specific metrics on the computational efficiency of our method in Tab. \ref{Computational Efficiency}. Our proposed models consistently achieve higher NDS and mAP scores than baseline BEVFusion \cite{liu2023bevfusion} while maintaining reasonable computational overhead. We attribute the speed improvement of our DepthFusion to our depth-aware feature fusion method. Due to the efficiency of our depth-aware feature fusion method, we can use a smaller backbone while maintaining competitive model performance. To further highlight the balance between performance and efficiency across our variants, we include a trade-off analysis in Figure \ref{trade-off}, which visualizes the relationship between NDS and both FLOPs and latency. Among all variants, DepthFusion-light achieves the most balanced trade-off, offering competitive performance with significantly lower computational cost and latency. In contrast, DepthFusion-huge delivers the best overall performance but requires higher FLOPs and latency, making it more suitable for offline or high-performance computing applications.

\subsection{Ablation Studies}
We conduct ablation studies to first demonstrate the effect of each component of DepthFusion, then to demonstrate the effect of depth encoding in DGF and DLF, and finally to assess the impact of multiplying depth encoding. All method variants are implemented on the nuScenes validation dataset.

\textbf{Effect of DGF and DLF.} To demonstrate the effect of DGF and DLF, we conduct experiments by integrating the components one by one into the baseline, BEVFusion \cite{liu2023bevfusion}. The results are shown in Tab. \ref{ablation-each module}. We find that our DGF improves the baseline performance by 1.0 pp w.r.t. NDS and 0.9 pp w.r.t. mAP. This demonstrates that dynamically adjusting the weights of the image BEV features during fusion is effective for 3D object detection. Additionally, our DLF improves the baseline performance by 1.3 pp w.r.t. NDS and 0.8 pp w.r.t. mAP, which indicates that dynamically adjusting the weights of the local raw instance features based on depth during fusion effectively compensates for the information loss caused by the transformation of global features into the BEV feature space. We also investigate the impact of the integration order. The results show that applying DGF before DLF yields better performance (69.8\% mAP vs. 69.0\% mAP). We analyze that DGF enhances the global BEV representation, providing a robust and spatially consistent scene-level context. When this enriched representation is established first, DLF can effectively refine local object features. In contrast, if DLF is applied before the global BEV features are sufficiently enhanced, its localized refinement may rely on suboptimal representations, leading to suboptimal detection results. This indicates that establishing a strong global foundation before local enhancement better aligns with the hierarchical nature of detection tasks.

    \begin{table}[t!]
        \centering
        \caption{Ablation studies of depth encoding ($D$) in DGF and DLF.}
        \centering
        \setlength{\tabcolsep}{1.7mm}{
        \scalebox{1.0}{
        \begin{tabular}{ c| c c} 
        \hline
        Methods & NDS & mAP \\
        \hline
        Baseline + DGF & 72.4 & 69.4 \\
        w/o $D$ & \quad \ \ 71.8\textcolor{red}{$^{\downarrow0.6}$} & \quad \ \ 69.0\textcolor{red}{$^{\downarrow0.4}$} \\
        \hline
        \hline
        Baseline + DLF & 72.7 & 69.3 \\
        w/o $D$ & \quad \ \ 71.6\textcolor{red}{$^{\downarrow1.1}$} & \quad \ \ 68.4\textcolor{red}{$^{\downarrow0.9}$} \\
        \hline
        \hline
        Baseline + DGF + DLF & 73.3 & 69.8 \\
         w/o $D$ & \quad \ \ 71.8\textcolor{red}{$^{\downarrow1.5}$} & \quad \ \ 68.6\textcolor{red}{$^{\downarrow1.2}$} \\
        \hline
        \end{tabular}
        }
        }
        \label{ablation-depth encoding position}
    \end{table}
    \begin{table}[t!]
        \centering
        \caption{Ablation studies of different embedding operations for depth encoding.}
        \centering
        \setlength{\tabcolsep}{5mm}{
        \scalebox{1.0}{
        \begin{tabular}{c| c c} 
        \hline
        Embedding Operations & NDS & mAP \\
        \hline
        Summation & 72.8 & 69.2 \\
        Concatenation & 72.5 & 68.7 \\
        Multiplication & \textbf{73.3} & \textbf{69.8} \\
        \hline
        \end{tabular}
        }
        }
        \label{ablation-depth encoding integrate method}
    \end{table}

    \begin{table}[t!]
        \centering
        \caption{Experiments of alternative depth encoding methods in DepthFusion-light.}
        \label{alternative depth encoding}
        \centering
        \setlength{\tabcolsep}{1.3mm}{
        \scalebox{1}{
        \begin{tabular}{ c| c c | c} 
        \hline
        Depth Encoding & NDS & mAP & Parameters (M) \\
        \hline
        Sine and cosine functions & \textbf{73.3} & \textbf{69.8} & - \\
        Parameterization & 71.8 & 67.8 & 8.3 \\
        \hline
        \end{tabular}
            }
        } 
    \end{table}
    
    \begin{figure}[t]
        \centering
        \includegraphics[width=0.91 \linewidth]{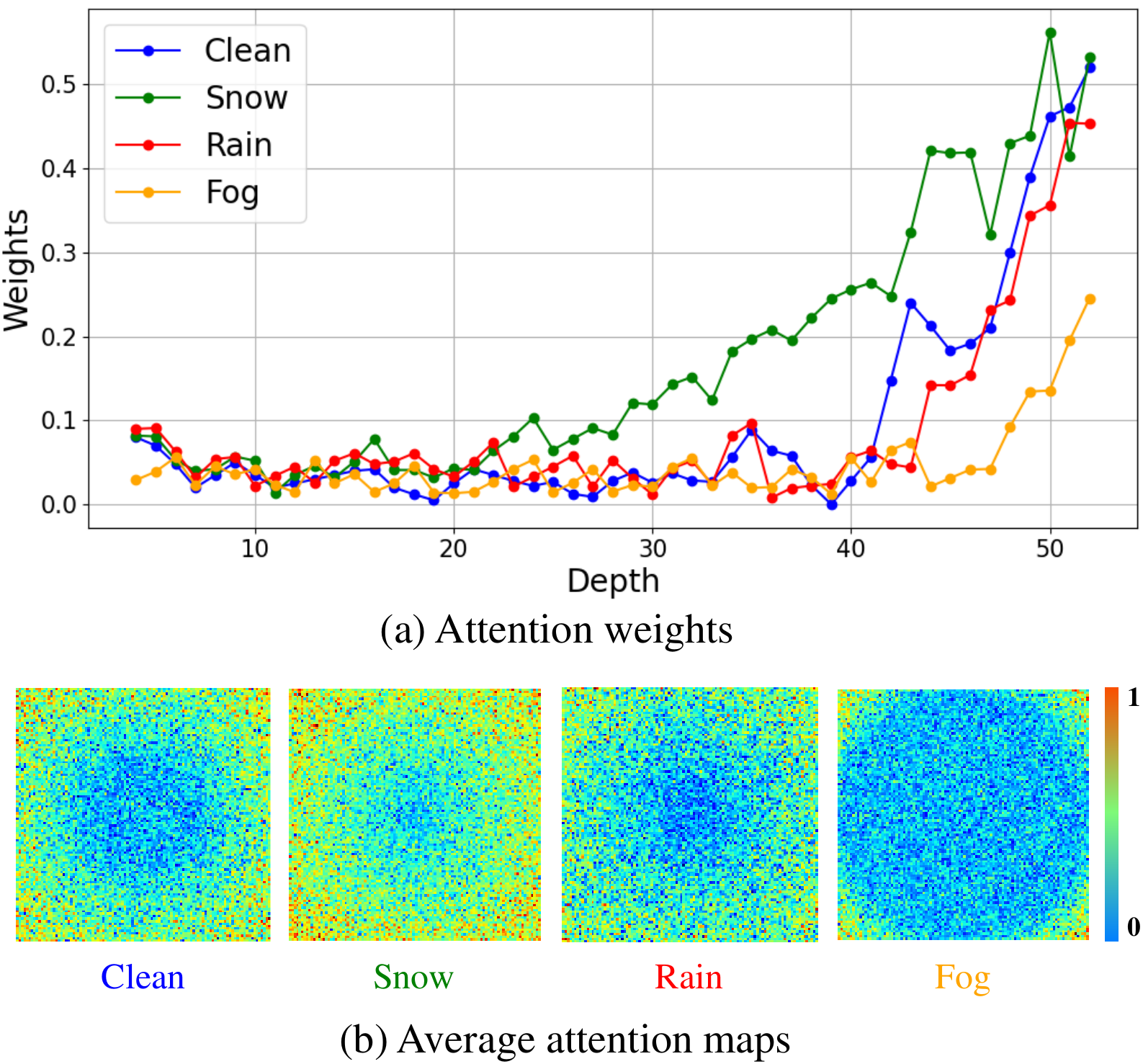}
        \caption{Statistics of attention weights and the visualization of attention maps under different weather conditions.}
        \label{weights_and_variation_figure}
    \end{figure}
    
\textbf{Effect of depth encoding in DGF and DLF.} To evaluate the effectiveness of our depth encoding, we conduct experiments where the depth encoding is removed from the DGF and DLF modules, respectively. The results are shown in Tab. \ref{ablation-depth encoding position}. When removing the depth encoding from Baseline+DGF, the performance drops by 0.6 pp w.r.t. NDS and 0.4 pp w.r.t. mAP. Similarly, when removing the depth encoding from Baseline+DLF, the performance also decreases by 1.1 pp w.r.t. NDS and 0.9 pp w.r.t. mAP. Additionally, when removing the depth encoding of both DGF and DLF, the performance also decreases by 1.5 pp w.r.t. NDS and 1.2 pp w.r.t. mAP. These results indicate that our depth encoding is effective. Furthermore, we observe that removing the depth encoding from the DLF module results in a larger performance drop, suggesting that depth encoding plays a more crucial role in local feature fusion. We analyze that local feature fusion theoretically relies on the fine-grained geometric details of objects to enhance feature representations. For instance, objects located at extreme distances often have sparse and incomplete point cloud representations, which lack sufficient fine-grained information. If the point cloud and image features of such an object are fused into the BEV representation with equal weights, this could degrade the BEV features and badly affect detection performance. Our depth encoding addresses this issue by dynamically reducing the weight of incomplete point cloud features, thereby mitigating the negative impact of incomplete point cloud features on BEV features. Simultaneously, it integrates the fine-grained information from images into the BEV features, achieving feature enhancement.

\textbf{Impact of different embedding operations for depth encoding.} We conduct experiments with different embedding operations of depth encoding, including concatenation, summation, and multiplication. The results in Tab. \ref{ablation-depth encoding integrate method}, show that the multiplication operation consistently outperforms the summation and concatenation operations w.r.t. both metrics. The superior performance of multiplication can be attributed to its ability to more effectively modulate the feature maps based on depth information. Unlike summation, which simply shifts the feature values, or concatenation, which increases the dimensionality without direct interaction, multiplication allows for more interaction between the depth encoding and features, leading to better feature representation and ultimately improving the detection performance.

\textbf{Effect of depth encoding.}
To explore alternative strategies for depth encoding, we experimented with a learnable matrix-based encoding. However, as shown in Tab. \ref{alternative depth encoding}, the parameterized encodings do not improve detection performance and even slightly degraded it in some cases, while also introducing over 8M additional parameters. We attribute this to the fact that parameterized encodings may obscure the true depth distribution, making it harder for the model to extract meaningful spatial cues. In contrast, our sine-cosine encoding is parameter-free, preserves the geometric depth structure in BEV space, and leads to more stable and efficient fusion.
    
\subsection{Visualization}
To better understand how depth encoding affects the feature fusion, in Fig. \ref{weights_and_variation_figure}, we plot a curve to observe how the attention weights applied to the image BEV features in our DGF module vary with depth, and visualize the average attention map under different weather conditions. First, for clean weather, it is evident that the weights of the image BEV features stay low in the near range, but go up significantly as depth increases when the depth is larger than 40 meters. This trend supports our hypothesis that the image modality would become more important as depth increases. In this way, our depth encoding allows the model to dynamically adjust the weights of image BEV features based on depth.

Next, for other weather conditions, we attribute the robustness of our methods to the combined effects of depth encoding. We analyze the reason for this using the foggy condition as an example: It is evident that fog has a greater impact on the image modality. Our depth-aware feature fusion adjusts the fusion weights based on the quality of the features. In such a case, the degradation of image feature quality leads to a reduction in the weight assigned to the image modality during fusion (orange line in Fig. \ref{weights_and_variation_figure} (a)). As a result, the fused features with depth guidance are of higher quality than those without depth guidance. In this way, it reduces the negative impact of corruption to some extent.
    
We also compare the detection results of our DepthFusion method with the baseline BEVFusion \cite{liu2023bevfusion} in Fig. \ref{vis_figure}, where we clearly find that our method better localizes those distant objects compared to BEVFusion. These results demonstrate that our proposed multi-modal fusion strategy based on depth is more effective for detection. Besides, we exhibit the corresponding BEV feature maps, where our method shows a stronger feature response for the foreground objects, especially for distant ones. That is why our feature fusion strategy can provide higher-quality detection results. More qualitative results can be found in supplementary materials.

\section{Conclusion}
In this paper, we are the first to observe that different modalities play different roles as depth varies via statistical analysis and visualization. Based on this finding, we propose a feature fusion strategy for LiDAR-camera 3D object detection, namely Depth-Aware Hybrid Feature Fusion (DepthFusion), that dynamically adjusts the weights of features during feature fusion by introducing depth encoding at both global and local levels. Extensive experiments on the nuScenes and KITTI datasets demonstrate that our DepthFusion method surpasses previous state-of-the-art methods. Moreover, our DepthFusion is more robust to various kinds of corruptions, outperforming previous methods on the nuScenes-C dataset. We hope our method offers useful insights for feature fusion in the field of LiDAR-camera 3D object detection.

\bibliographystyle{plain}
\bibliography{reference}

\begin{thebibliography}{10}

\bibitem{caesar2020nuscenes}
Holger Caesar, Varun Bankiti, Alex~H Lang, Sourabh Vora, Venice~Erin Liong, Qiang Xu, Anush Krishnan, Yu~Pan, Giancarlo Baldan, and Oscar Beijbom.
\newblock nuscenes: A multimodal dataset for autonomous driving.
\newblock In {\em Conference on Computer Vision and Pattern Recognition}, pages 11621--11631, 2020.

\bibitem{geiger2012kitti}
Andreas Geiger, Philip Lenz, and Raquel Urtasun.
\newblock Are we ready for autonomous driving? the kitti vision benchmark suite.
\newblock In {\em Conference on Computer Vision and Pattern Recognition}, pages 3354--3361, 2012.

\bibitem{dong2023nuscenes-c}
Yinpeng Dong, Caixin Kang, Jinlai Zhang, Zijian Zhu, Yikai Wang, Xiao Yang, Hang Su, Xingxing Wei, and Jun Zhu.
\newblock Benchmarking robustness of 3d object detection to common corruptions.
\newblock In {\em Conference on Computer Vision and Pattern Recognition}, pages 1022--1032, 2023.

\bibitem{shi2019pointrcnn}
Shaoshuai Shi, Xiaogang Wang, and Hongsheng Li.
\newblock Pointrcnn: 3d object proposal generation and detection from point cloud.
\newblock In {\em Conference on Computer Vision and Pattern Recognition}, pages 770--779, 2019.

\bibitem{zhou2018voxelnet}
Yin Zhou and Oncel Tuzel.
\newblock Voxelnet: End-to-end learning for point cloud based 3d object detection.
\newblock In {\em Proceedings of the IEEE Conference on Computer Vision and Pattern Recognition}, pages 4490--4499, 2018.

\bibitem{an2023sp}
Pei An, Yucong Duan, Yuliang Huang, Jie Ma, Yanfei Chen, Liheng Wang, You Yang, and Qiong Liu.
\newblock Sp-det: Leveraging saliency prediction for voxel-based 3d object detection in sparse point cloud.
\newblock {\em IEEE Transactions on Multimedia}, 2023.

\bibitem{shi2020pv}
Shaoshuai Shi, Chaoxu Guo, Li~Jiang, Zhe Wang, Jianping Shi, Xiaogang Wang, and Hongsheng Li.
\newblock Pv-rcnn: Point-voxel feature set abstraction for 3d object detection.
\newblock In {\em Conference on Computer Vision and Pattern Recognition}, pages 10529--10538, 2020.

\bibitem{wang2021fcos3d}
Tai Wang, Xinge Zhu, Jiangmiao Pang, and Dahua Lin.
\newblock Fcos3d: Fully convolutional one-stage monocular 3d object detection.
\newblock In {\em International Conference on Computer Vision}, pages 913--922, 2021.

\bibitem{huang2021bevdet}
Junjie Huang, Guan Huang, Zheng Zhu, Yun Ye, and Dalong Du.
\newblock Bevdet: High-performance multi-camera 3d object detection in bird-eye-view.
\newblock {\em arXiv preprint arXiv:2112.11790}, 2021.

\bibitem{vora2020pointpainting}
Sourabh Vora, Alex~H Lang, Bassam Helou, and Oscar Beijbom.
\newblock Pointpainting: Sequential fusion for 3d object detection.
\newblock In {\em Conference on Computer Vision and Pattern Recognition}, pages 4604--4612, 2020.

\bibitem{wang2021pointaugmenting}
Chunwei Wang, Chao Ma, Ming Zhu, and Xiaokang Yang.
\newblock Pointaugmenting: Cross-modal augmentation for 3d object detection.
\newblock In {\em Conference on Computer Vision and Pattern Recognition}, pages 11794--11803, 2021.

\bibitem{wu2023virtual}
Hai Wu, Chenglu Wen, Shaoshuai Shi, Xin Li, and Cheng Wang.
\newblock Virtual sparse convolution for multimodal 3d object detection.
\newblock In {\em Conference on Computer Vision and Pattern Recognition}, pages 21653--21662, 2023.

\bibitem{wang2024virpnet}
Lin Wang, Shiliang Sun, and Jing Zhao.
\newblock Virpnet: A multimodal virtual point generation network for 3d object detection.
\newblock {\em IEEE Transactions on Multimedia}, 2024.

\bibitem{bai2022transfusion}
Xuyang Bai, Zeyu Hu, Xinge Zhu, Qingqiu Huang, Yilun Chen, Hongbo Fu, and Chiew-Lan Tai.
\newblock Transfusion: Robust lidar-camera fusion for 3d object detection with transformers.
\newblock In {\em Conference on Computer Vision and Pattern Recognition}, pages 1090--1099, 2022.

\bibitem{zhu2022vpfnet}
Hanqi Zhu, Jiajun Deng, Yu~Zhang, Jianmin Ji, Qiuyu Mao, Houqiang Li, and Yanyong Zhang.
\newblock Vpfnet: Improving 3d object detection with virtual point based lidar and stereo data fusion.
\newblock {\em IEEE Transactions on Multimedia}, 2022.

\bibitem{chen2023futr3d}
Xuanyao Chen, Tianyuan Zhang, Yue Wang, Yilun Wang, and Hang Zhao.
\newblock Futr3d: A unified sensor fusion framework for 3d detection.
\newblock In {\em Conference on Computer Vision and Pattern Recognition}, pages 172--181, 2023.

\bibitem{zhu2020deformable}
Xizhou Zhu, Weijie Su, Lewei Lu, Bin Li, Xiaogang Wang, and Jifeng Dai.
\newblock Deformable detr: Deformable transformers for end-to-end object detection.
\newblock {\em arXiv preprint arXiv:2010.04159}, 2020.

\bibitem{lin20223d}
Chunmian Lin, Daxin Tian, Xuting Duan, Jianshan Zhou, Dezong Zhao, and Dongpu Cao.
\newblock 3d-dfm: Anchor-free multimodal 3-d object detection with dynamic fusion module for autonomous driving.
\newblock {\em IEEE Transactions on Neural Networks and Learning Systems}, 34(12):10812--10822, 2022.

\bibitem{lin2022cl3d}
Chunmian Lin, Daxin Tian, Xuting Duan, Jianshan Zhou, Dezong Zhao, and Dongpu Cao.
\newblock Cl3d: Camera-lidar 3d object detection with point feature enhancement and point-guided fusion.
\newblock {\em IEEE Transactions on Intelligent Transportation Systems}, 23(10):18040--18050, 2022.

\bibitem{liang2022bevfusion}
Tingting Liang, Hongwei Xie, Kaicheng Yu, Zhongyu Xia, Zhiwei Lin, Yongtao Wang, Tao Tang, Bing Wang, and Zhi Tang.
\newblock Bevfusion: A simple and robust lidar-camera fusion framework.
\newblock {\em Advances in Neural Information Processing Systems}, 35:10421--10434, 2022.

\bibitem{liu2023bevfusion}
Zhijian Liu, Haotian Tang, Alexander Amini, Xinyu Yang, Huizi Mao, Daniela~L Rus, and Song Han.
\newblock Bevfusion: Multi-task multi-sensor fusion with unified bird's-eye view representation.
\newblock In {\em 2023 IEEE International Conference on Robotics and Automation}, pages 2774--2781. IEEE, 2023.

\bibitem{xie2023sparsefusion}
Yichen Xie, Chenfeng Xu, Marie-Julie Rakotosaona, Patrick Rim, Federico Tombari, Kurt Keutzer, Masayoshi Tomizuka, and Wei Zhan.
\newblock Sparsefusion: Fusing multi-modal sparse representations for multi-sensor 3d object detection.
\newblock In {\em International Conference on Computer Vision}, pages 17591--17602, 2023.

\bibitem{cai2023objectfusion}
Qi~Cai, Yingwei Pan, Ting Yao, Chong-Wah Ngo, and Tao Mei.
\newblock Objectfusion: Multi-modal 3d object detection with object-centric fusion.
\newblock In {\em International Conference on Computer Vision}, pages 18067--18076, 2023.

\bibitem{yan2023cmt}
Junjie Yan, Yingfei Liu, Jianjian Sun, Fan Jia, Shuailin Li, Tiancai Wang, and Xiangyu Zhang.
\newblock Cross modal transformer via coordinates encoding for 3d object dectection.
\newblock {\em arXiv preprint arXiv:2301.01283}, 2023.

\bibitem{li2023logonet}
Xin Li, Tao Ma, Yuenan Hou, Botian Shi, Yuchen Yang, Youquan Liu, Xingjiao Wu, Qin Chen, Yikang Li, Yu~Qiao, et~al.
\newblock Logonet: Towards accurate 3d object detection with local-to-global cross-modal fusion.
\newblock In {\em Conference on Computer Vision and Pattern Recognition}, pages 17524--17534, 2023.

\bibitem{yin2024isfusion}
Junbo Yin, Jianbing Shen, Runnan Chen, Wei Li, Ruigang Yang, Pascal Frossard, and Wenguan Wang.
\newblock Is-fusion: Instance-scene collaborative fusion for multimodal 3d object detection.
\newblock In {\em Conference on Computer Vision and Pattern Recognition}, pages 14905--14915, 2024.

\bibitem{pang2020clocs}
Su~Pang, Daniel Morris, and Hayder Radha.
\newblock Clocs: Camera-lidar object candidates fusion for 3d object detection.
\newblock In {\em 2020 IEEE/RSJ International Conference on Intelligent Robots and Systems}, pages 10386--10393. IEEE, 2020.

\bibitem{vaswani2017attention}
Ashish Vaswani, Noam Shazeer, Niki Parmar, Jakob Uszkoreit, Llion Jones, Aidan~N Gomez, {\L}ukasz Kaiser, and Illia Polosukhin.
\newblock Attention is all you need.
\newblock {\em Advances in neural information processing systems}, 30, 2017.

\bibitem{deng2021voxel}
Jiajun Deng, Shaoshuai Shi, Peiwei Li, Wengang Zhou, Yanyong Zhang, and Houqiang Li.
\newblock Voxel r-cnn: Towards high performance voxel-based 3d object detection.
\newblock In {\em Proceedings of the AAAI Conference on Artificial Intelligence}, pages 1201--1209, 2021.

\bibitem{he2017mask}
Kaiming He, Georgia Gkioxari, Piotr Dollar, and Ross Girshick.
\newblock Mask r-cnn.
\newblock {\em IEEE transactions on pattern analysis and machine intelligence}, 42(2):386--397, 2020.

\bibitem{mmdet3d2020}
MMDetection3D Contributors.
\newblock {MMDetection3D: OpenMMLab} next-generation platform for general {3D} object detection.
\newblock \url{https://github.com/open-mmlab/mmdetection3d}, 2020.

\bibitem{yan2018second}
Yan Yan, Yuxing Mao, and Bo~Li.
\newblock Second: Sparsely embedded convolutional detection.
\newblock {\em Sensors}, 18(10):3337, 2018.

\bibitem{he2016resnet}
Kaiming He, Xiangyu Zhang, Shaoqing Ren, and Jian Sun.
\newblock Deep residual learning for image recognition.
\newblock In {\em Proceedings of the IEEE Conference on Computer Vision and Pattern Recognition}, pages 770--778, 2016.

\bibitem{liu2022convnet}
Zhuang Liu, Hanzi Mao, Chao-Yuan Wu, Christoph Feichtenhofer, Trevor Darrell, and Saining Xie.
\newblock A convnet for the 2020s.
\newblock In {\em Conference on Computer Vision and Pattern Recognition}, pages 11976--11986, 2022.

\bibitem{lin2017fpn}
Tsung-Yi Lin, Piotr Doll{\'a}r, Ross Girshick, Kaiming He, Bharath Hariharan, and Serge Belongie.
\newblock Feature pyramid networks for object detection.
\newblock In {\em Proceedings of the IEEE Conference on Computer Vision and Pattern Recognition}, pages 2117--2125, 2017.

\bibitem{huang2022bevpoolv2}
Junjie Huang and Guan Huang.
\newblock Bevpoolv2: A cutting-edge implementation of bevdet toward deployment.
\newblock {\em arXiv:2211.17111}, 2022.

\bibitem{yang2022deepinteraction}
Zeyu Yang, Jiaqi Chen, Zhenwei Miao, Wei Li, Xiatian Zhu, and Li~Zhang.
\newblock Deepinteraction: 3d object detection via modality interaction.
\newblock {\em Advances in Neural Information Processing Systems}, 35:1992--2005, 2022.

\bibitem{jiao2023msmdfusion}
Yang Jiao, Zequn Jie, Shaoxiang Chen, Jingjing Chen, Lin Ma, and Yu-Gang Jiang.
\newblock Msmdfusion: Fusing lidar and camera at multiple scales with multi-depth seeds for 3d object detection.
\newblock In {\em Conference on Computer Vision and Pattern Recognition}, pages 21643--21652, 2023.

\bibitem{chen2023focalformer3d}
Yilun Chen, Zhiding Yu, Yukang Chen, Shiyi Lan, Anima Anandkumar, Jiaya Jia, and Jose~M Alvarez.
\newblock Focalformer3d: focusing on hard instance for 3d object detection.
\newblock In {\em International Conference on Computer Vision}, pages 8394--8405, 2023.

\bibitem{liu2021swintransformer}
Ze~Liu, Yutong Lin, Yue Cao, Han Hu, Yixuan Wei, Zheng Zhang, Stephen Lin, and Baining Guo.
\newblock Swin transformer: Hierarchical vision transformer using shifted windows.
\newblock In {\em International Conference on Computer Vision}, pages 10012--10022, 2021.

\bibitem{li2024gafusion}
Xiaotian Li, Baojie Fan, Jiandong Tian, and Huijie Fan.
\newblock Gafusion: Adaptive fusing lidar and camera with multiple guidance for 3d object detection.
\newblock In {\em Conference on Computer Vision and Pattern Recognition}, pages 21209--21218, 2024.

\bibitem{chen2022AutoAlignV2}
Zehui Chen, Zhenyu Li, Shiquan Zhang, Liangji Fang, Qinhong Jiang, and Feng Zhao.
\newblock Deformable feature aggregation for dynamic multi-modal 3d object detection.
\newblock In {\em Proceedings of the European Conference on Computer Vision}, pages 628--644. Springer, 2022.

\bibitem{wang2020cspnet}
Chien-Yao Wang, Hong-Yuan~Mark Liao, Yueh-Hua Wu, Ping-Yang Chen, Jun-Wei Hsieh, and I-Hau Yeh.
\newblock Cspnet: A new backbone that can enhance learning capability of cnn.
\newblock In {\em Conference on Computer Vision and Pattern Recognition Workshops}, pages 390--391, 2020.

\bibitem{li2022unifying}
Yanwei Li, Yilun Chen, Xiaojuan Qi, Zeming Li, Jian Sun, and Jiaya Jia.
\newblock Unifying voxel-based representation with transformer for 3d object detection.
\newblock {\em Advances in Neural Information Processing Systems}, 35:18442--18455, 2022.

\bibitem{lee2019vovnet}
Youngwan Lee, Joong-won Hwang, Sangrok Lee, Yuseok Bae, and Jongyoul Park.
\newblock An energy and gpu-computation efficient backbone network for real-time object detection.
\newblock In {\em Conference on Computer Vision and Pattern Recognition Workshops}, pages 0--0, 2019.

\bibitem{wang2023unitr}
Haiyang Wang, Hao Tang, Shaoshuai Shi, Aoxue Li, Zhenguo Li, Bernt Schiele, and Liwei Wang.
\newblock Unitr: A unified and efficient multi-modal transformer for bird's-eye-view representation.
\newblock In {\em International Conference on Computer Vision}, pages 6792--6802, 2023.

\bibitem{wang2023dsvt}
Haiyang Wang, Chen Shi, Shaoshuai Shi, Meng Lei, Sen Wang, Di~He, Bernt Schiele, and Liwei Wang.
\newblock Dsvt: Dynamic sparse voxel transformer with rotated sets.
\newblock In {\em Conference on Computer Vision and Pattern Recognition}, pages 13520--13529, 2023.

\bibitem{yang2024unipad}
Honghui Yang, Sha Zhang, Di~Huang, Xiaoyang Wu, Haoyi Zhu, Tong He, Shixiang Tang, Hengshuang Zhao, Qibo Qiu, Binbin Lin, et~al.
\newblock Unipad: A universal pre-training paradigm for autonomous driving.
\newblock In {\em Conference on Computer Vision and Pattern Recognition}, pages 15238--15250, 2024.

\bibitem{zhang2024sparselif}
Hongcheng Zhang, Liu Liang, Pengxin Zeng, Xiao Song, and Zhe Wang.
\newblock Sparselif: High-performance sparse lidar-camera fusion for 3d object detection.
\newblock In {\em Proceedings of the European Conference on Computer Vision}, pages 109--128. Springer, 2024.

\bibitem{xu2018pointfusion}
Danfei Xu, Dragomir Anguelov, and Ashesh Jain.
\newblock Pointfusion: Deep sensor fusion for 3d bounding box estimation.
\newblock In {\em Proceedings of the IEEE Conference on Computer Vision and Pattern Recognition}, pages 244--253, 2018.

\bibitem{huang2020epnet}
Tengteng Huang, Zhe Liu, Xiwu Chen, and Xiang Bai.
\newblock Epnet: Enhancing point features with image semantics for 3d object detection.
\newblock In {\em Proceedings of the European Conference on Computer Vision}, pages 35--52. Springer, 2020.

\bibitem{zhang2022cat}
Yanan Zhang, Jiaxin Chen, and Di~Huang.
\newblock Cat-det: Contrastively augmented transformer for multi-modal 3d object detection.
\newblock In {\em Conference on Computer Vision and Pattern Recognition}, pages 908--917, 2022.

\bibitem{chen2022focal}
Yukang Chen, Yanwei Li, Xiangyu Zhang, Jian Sun, and Jiaya Jia.
\newblock Focal sparse convolutional networks for 3d object detection.
\newblock In {\em Conference on Computer Vision and Pattern Recognition}, pages 5428--5437, 2022.

\bibitem{li2022voxel}
Yanwei Li, Xiaojuan Qi, Yukang Chen, Liwei Wang, Zeming Li, Jian Sun, and Jiaya Jia.
\newblock Voxel field fusion for 3d object detection.
\newblock In {\em Conference on Computer Vision and Pattern Recognition}, pages 1120--1129, 2022.

\end{thebibliography}



\vfill

\end{document}